\documentclass{article}
\usepackage{spconf,amsmath,graphicx}

\title{Retrieval-Augmented Natural Language Reasoning for Explainable Visual Question Answering}
\name{Su Hyeon Lim, Minkuk Kim, Hyeon Bae Kim, Seong Tae Kim\textsuperscript{*}\thanks{This work was supported in part by the National Research Foundation of Korea(NRF) grant funded by the Korea government(MSIT) (No. RS-2024-00334321), by the Institute of Information and Communications Technology Planning and Evaluation (IITP) Grant funded by the Korea Government (MSIT) under Grant 2022-0-00078 (Explainable Logical Reasoning for Medical Knowledge Generation), and by Center for Applied Research in Artificial Intelligence (CARAI) grant funded by DAPA and ADD (UD230017TD). \textsuperscript{*}Dr. Seong Tae Kim is a corresponding author.}}
\address{Augmented Intelligence Lab. Kyung Hee University, Republic of Korea}

\begin{document}
%
\maketitle
\begin{abstract}


Visual Question Answering with Natural Language Explanation (VQA-NLE) task is challenging due to its high demand for reasoning-based inference.
Recent VQA-NLE studies focus on enhancing model networks
to amplify the model’s reasoning capability but this approach is resource-consuming and unstable.
In this work, we introduce a new VQA-NLE model, ReRe (Retrieval-augmented natural language Reasoning), using leverage retrieval information from the memory to aid in generating accurate answers and persuasive explanations without relying on complex networks and extra datasets.
ReRe is an encoder-decoder architecture model using a pre-trained clip vision encoder and a pre-trained GPT-2 language model as a decoder.
Cross-attention layers are added in the GPT-2 for processing retrieval feature.
ReRe outperforms previous methods in VQA accuracy and explanation score and shows improvement in NLE with more persuasive, reliability.

\end{abstract}
\section{Introduction}
\label{sec:intro}

With significant advancements in deep learning models, there has been notable progress in vision-language tasks such as image captioning and visual question answering (VQA).
To substantiate the performance improvements in vision-language (VL) tasks, explanation has become crucial.
The importance of natural language explanations (NLE) is further emphasized, particularly for applying vision-language tasks based on principles of truth, correctness, and understanding \cite{anderson2018vision,feng2023nle}. However, NLE for vision and VL tasks remains a challenging task due to its high demand for reasoning-based inference \cite{RATIONALETRANSFORMER}.
It requires both an understanding of the image and a higher-level reasoning ability beyond VQA to prove the answers.
With a lack of reasoning, models generate explanations that are non-relate to the predicted answers or are completely wrong.

Recent NLE studies focus on enhancing model networks to amplify the model’s reasoning capability \cite{lai2024towards, suo2023s3c,ge2023wrong}. This approach is ideal for obtaining high reasoning ability without relying on large-scale model architecture and data, efficiently producing explanations through their unique logical processes \cite{lu2022learn}.
However, These unique logical processes generally consist of many reasoning steps and are often recursive.
Constructing such complex networks is resource-consuming and in some cases, results get worse over steps \cite{ge2023wrong}.

In this work, we introduce a new VQA-NLE model, ReRe (\textbf{Re}trieval-augmented natural language \textbf{Re}asoning), using leverage retrieval information from the memory to aid in generating accurate answers and persuasive explanations without relying on complex networks and extra datasets.
Recent research has demonstrated significant results by applying retrieval augmentation to various vision-language tasks such as video question answering \cite{pan2023retrieving}, image captioning \cite{ramos2023smallcap,sarto2022retrieval}.
These studies show that by providing semantic features from retrieval augmentation to the model, the model performance can be enhanced.
Inspired by this approach, we design a retrieval augmentation to the NLE task, specifically for the VQA-NLE task.
Our model uses a pre-trained CLIP vision encoder and a pre-trained GPT-2 language model as a decoder.
A new layer is added in GPT-2 to perform cross-attention over the encoded retrieval features, which is an extremely simple way to gain reasoning from retrieval features.
After retrieval of the memory database, semantic retrieval features are extracted by averaging the encoded sample feature.
These retrieval features are then inputted with the image feature and question feature encoded with the clip vision encoder.
ReRe generates the answer and explanation from the given image and question with the aid of retrieval features.
Compared with other methods, ReRe shows improvement in explanation with more persuasive, reliability.

\section{Related Work}
\label{sec:format}

{\bf VQA-NLE:}
VQA has firstly proposed by \cite{firstVQA} that answering questions about the given real-world images.
Since then, many approaches have been proposed on VQA task \cite{vqa1,vqa3}.
To pursue explainable VQA with reasoning process, NLE task has been proposed in \cite{first-vqa-nle}. Textual explanation of classification decision is generated for end-user, which is different from lower-level explanations that apply visualization technologies\cite{gradcam}.
\cite{vqax} proposed VQA-X datasets and PJ-X model.
VQA-X gives a rational explanation of visual question answering tasks.
PJ-X consists of an answering model and a multimodal explanation model, in which the predicted answer of the answering model is used to generate textual justifications in the explanation model.
e-ug \cite{e-ug} model is also separated from VL-model (UNITER \cite{uniter}) that predicts answers and pre-trained language model (GPT-2 \cite{gpt2}) to generate explanation.
\cite{RATIONALETRANSFORMER} suggests a model architecture that generates text explanation by GPT-2’s backbone architecture. 
NLX-GPT \cite{sammani2022nlx} is a unified model that simultaneously generates answers and explanations. By unifying the VL model and explanation model in one, their answer and explanation are more correlated.

Recently, $S_{3}C$ \cite{suo2023s3c} used self-critical learning networks to improve the model's self-interpretability.
MCLE \cite{whitehouse2023towards} improves NLE ability by chain-of-thought strategy in generating explanation and multi-level contrastive learning network.
ReVisE \cite{ge2023wrong} introduces recursive networks where the generated explanation is utilized for next-step explanation generation.

{\bf Retrieval Augmentation:}
Retrieval augmentation has gained attention in natural language processing (NLP) \cite{retrieval1,retrieval2,retrieval4} and also various multimodal tasks such as image and video captioning \cite{ramos2023smallcap,kim2024you} and VQA\cite{pan2023retrieving}.
Despite of achievement of retrieval augmentation in many tasks, there is no attempt in NLE tasks. To the best of our knowledge, this is the first study to design retrieval augmentation for NLE tasks.

\begin{figure}[t!]
    \centering
    \centerline{\includegraphics[width=0.5\textwidth]{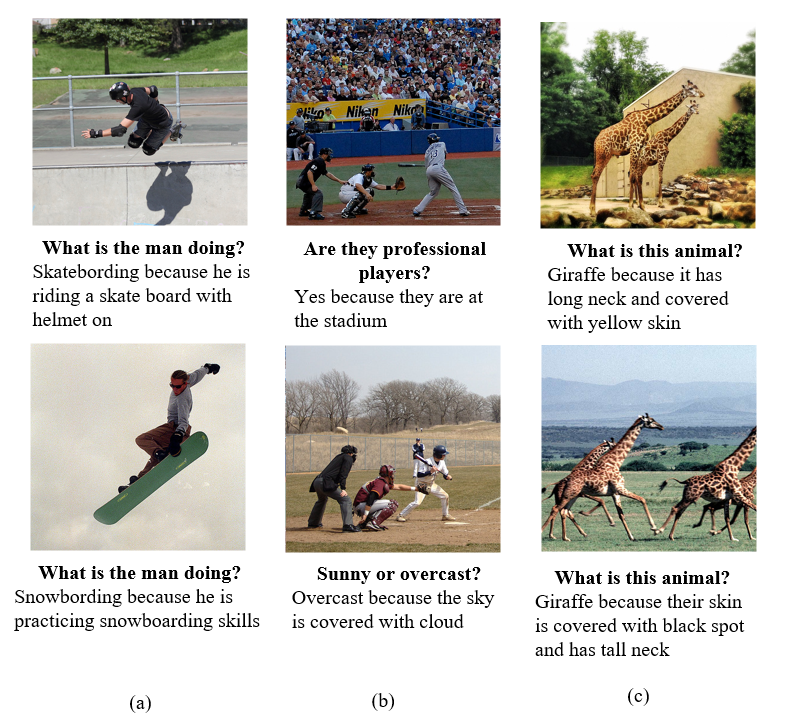}}
    \vspace{-0.3cm}
    \caption{Examples of retrieval in our method. (a) The question types between the query (left) and retrieval (right) samples are the same, but the images contain different semantic information, (b) the images contain the same semantic information, but the contents of the questions are different.
    (c) is an ideal case where both the image and the question contents have the same semantics.}
    \vspace{-0.1cm}
\end{figure}

\section{Method}
Our goal is to retrieve related samples from a memory database and properly utilize them to enhance the performance of VQA and NLE.
In subsection 3.1, we introduce how to retrieve informative samples from memory database. Our definition of an informative sample is a sample with the same question type and similar semantic information. Fig.1 shows examples of retrieval in our method.
In subsection 3.2, we introduce details of ReRe's architecture which processes input image, question, and retrieval features.

\subsection{Retrieval Method}

The VQA-NLE task is to generate an answer from the question based on visual information from the image, along with an explanation that justifies the answer.
To help the model generate more accurate and informative answers and explanations, we build a memory database that consists of images, questions, answers, and explanations from the training dataset.
To retrieve a sample, we use a question and an image of the input query which are used as a key for searching and retrieving answers and explanations from similar cases.

First, we measure the similarity score between the query question ($Q_q$) and the sample question ($Q_s$) using cosine similarity, and we also measure the similarity score between the query image ($I_q$) and the sample explanation ($E_s$).
To measure cosine similarity between text-text and text-image, we encoded text and images into feature representation using a pre-trained CLIP model.
Note that the CLIP model, trained with text and images in the same embedding space, can be used to measure similarity between multimodal features \cite{radford2021clip}.

Text-text cosine similarity gets higher when sentence structure is similar and identical words are used.
By measuring the similarity between the query question and the sample questions, we can retrieve samples that have the same question type (sentence structure is similar) and deal with the same domain (identical words are used).
Also, we measure the similarity between the query image and the sample explanation in order to retrieve a sample that contains semantic information relevant to the problem we want to solve.
The query's semantic information is contained in the image and the retrieved sample's semantic information is mainly extracted in the explanation.
By comparison with the retrieved sample's image and explanation, explanation information contains necessary image information in specific question situations.

To calculate the final similarity score between the query and the samples in the memory database, these two scores are combined as

\begin{equation}
Score_{Retrieval} = cos(Q_{q},Q_{s}) + cos(I_{q},E_{s}).
\end{equation}
After measuring the similarity score for all samples in the memory database with respect to the given query, our method retrieves the top-K samples based on their similarity scores.
In our work, we retrieve 10 samples (K=10) for each query.

\begin{figure}
    \centering
    \centerline{\includegraphics[width=0.5\textwidth]{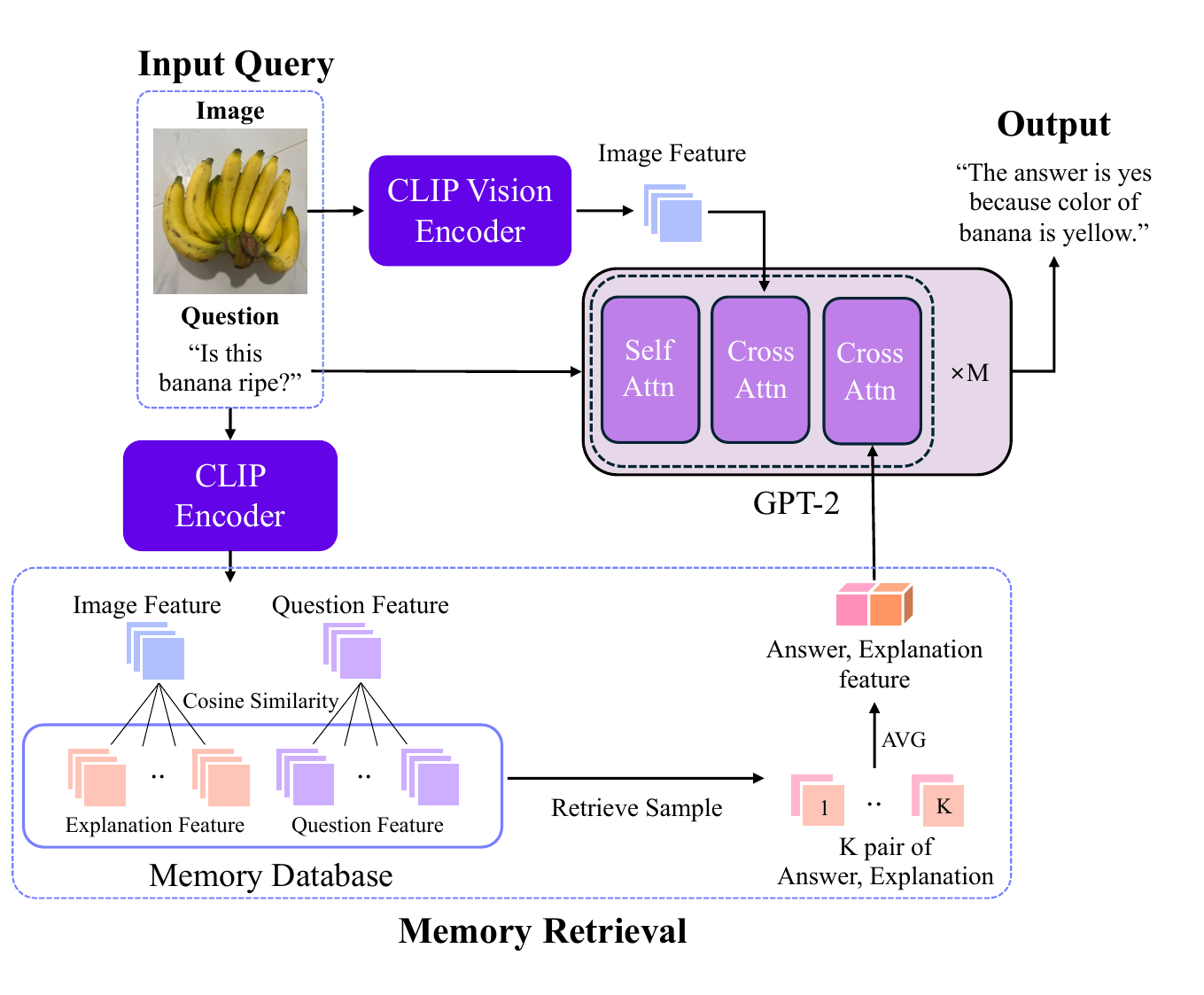}}
    \vspace{-0.4cm}
    \caption{Overall architecture of the proposed method. }
    \vspace{-0.1cm}
\end{figure}

After retrieving, we utilize answers and explanations of retrieval samples.
These K pairs of answers and explanations are encoded into features using the CLIP model, and the answer features and explanation features are each averaged to assist the model's reasoning.
By averaging K number of answers and explanations feature, it shows the effect that each feature representations concentrate on necessary representation.
Averaging features could refine noisy effects and reduce the computation complexity when they are used in the language model.

\subsection{Model}

{\bf Features Encoding:}
ReRe consists of an encoder-decoder architecture as depicted in Fig.2.
In the encoder part, a pre-trained CLIP model is used to encode the input image and retrieval text into feature representations.

The retrieval information to be inputted into the model from the retrieved samples consists of the answer and explanation.
The answer from the retrieved sample can serve as a hint to accurately answer the given query question.
Additionally, providing an explanation can help the model provide a logical justification for the answer it has given.

\noindent {\bf Features Cross Attention:}
ReRe's decoder part consists of pre-trained distilled GPT-2 and newly added cross-attention layers to deal with image, question, and retrieval features.
The original GPT's block structure consists of a self-attention layer and a cross-attention layer. 
In ReRe, the basic structure's self-attention layer processes the question text, and the cross-attention layer processes the image.
The newly added cross-attention layer follows the cross-attention layer that processes the image, and it handles retrieval features in that part.
In detail, hidden states obtained from the question feature embedding after passing through self-attention are cross-attention with the image features, output semantic cross-modal features that contain unified information about image and question\cite{ramos2023smallcap, sammani2022nlx}.
These semantic cross-modal features are then cross-attention with the retrieval features, and this process is repeated in every language model block.
Through these sequential attention computation procedures, the final answer to the query question is output based on the retrieval information.
This structure can incorporate retrieval information while preserving the general language capabilities of the original language model.

\noindent {\bf Answer Explanation Generation:}
Our model is a unified model that generates both the answer and explanation simultaneously. The model generates sentences based on retrieval-based inference, and these sentences are structured as ``\text{the answer is [answer] because [explanation]}".
Generating both the answer and explanation simultaneously in this way can increase the logical coherence between the answer and explanation \cite{sammani2022nlx}.

\section{Experiment}
\label{sec:pagestyle}

{\bf Dataset:}
Our experiments are conducted using VQA-X \cite{vqax}, which is widely used in the VQA-NLE task.
VQA-X is human annotated multimodal explanations for visual question answering.
It consists of 28K images, 33K Q/A pairs, and 42K explanations.
Out of the 33K Q/A pairs, 2.9K are used for training, 1.4K for validation, and 1.9K for testing.
The questions in VQA-X are composed of problems where answers and explanations need to be based on visual information, such as ``Is this banana ripe?".

\noindent {\bf Retrieval Memory Database:}
Our retrieval memory database is set up of VQA-X data.
During the training process, retrieval is conducted using only the training data from VQA-X in the Retrieval Memory Database.
During inference, the Database is constructed using both the training and validation data from VQA-X.
This utilization could set retrieval data to the fullest extent at the training and validation stage without any data leakage.
The retrieval samples in the Memory Database consist of images, question-answer pairs, and explanations, the same as the original VQA-X data configuration.

\noindent {\bf Training:}
The weights of the newly added cross-attention layers to the pre-trained distilled GPT-2 model were initialized randomly.
The Clip encoder is used only for extracting feature representations and is excluded from training.

\noindent {\bf Evaluation Metrics:}
We computed scores for prediction using automatic natural language generation (NLG) metrics including N-gram-based metrics BLEU-4 \cite{papineni2002bleu}, METEOR \cite{banerjee2005meteor}, ROUGE-L \cite{lin2004rouge}, and CIDEr \cite{vedantam2015cider}, as well as SPICE \cite{anderson2016spice} and BERTScore \cite{zhang2019bertscore}, which focus more on the semantic information of explanations.

\begin{table}[t]
\scalebox{0.85}{
\begin{tabular}{l|llllll|l}
\hline
        & B4   & M     & R    & C     & S    & BS    & Acc                    \\ \hline
PJ-X \cite{vqax}    & 22.7 & 19.7 & 46.0 & 82.7  & 17.1 & 84.6  & 76.4                   \\
FME \cite{FME}    & 23.1 & 20.4 & 47.1 & 87.0  & 18.4 & 85.2  & 75.5                   \\
RVT \cite{RATIONALETRANSFORMER}    & 17.4 & 19.2 & 42.1 & 52.5  & 15.8 & 85.7  & 68.6                   \\
e-UG \cite{e-ug}   & 23.2 & 22.1 & 45.7 & 49.9  & 20.1 & 87.0  & 80.5                   \\
NLX-GPT \cite{sammani2022nlx} & 28.5 & 23.1 & 51.5 & 110.6 & 22.1 & 86.9  & 83.1                  \\
ReVisE \cite{ge2023wrong} & 28.2 & 23.2 & 51.8 & 108.9 & 22.6 & 88.1  & \multicolumn{1}{c}{\_} \\ \hline
\textbf{ReRe$_I$}  & \textbf{28.7} & \textbf{23.4} & \textbf{52.0} & \textbf{111.7} & \textbf{22.7} & \textbf{90.1} & \textbf{83.0} \\
\textbf{ReRe} & \textbf{29.2} & \textbf{23.4} & \textbf{52.1} & \textbf{113.4} & \textbf{22.7} & \textbf{90.2} & \textbf{83.7} \\ \hline
\end{tabular}
}
\caption{Filtered Scores comparison with the state-of-the-art model on the VQA-X datasets. Filtered scores only consider the samples that have correct answers. B4, M, R, C, S, BS, Acc are short for BLEU-4 \cite{papineni2002bleu}, METEOR \cite{banerjee2005meteor}, ROUGE-L \cite{lin2004rouge}, CIDEr \cite{vedantam2015cider}, SPICE \cite{anderson2016spice}, BERTSCORE \cite{zhang2019bertscore}, answer accuracy.
ReRe$_I$ denotes the result of measuring retrieval score with a cosine similarity of image-image($cos(I_{q},I_{s})$), instead of the purposed method $cos(I_{q},E_{s})$.}
\label{tab:my-table}
\end{table}

\begin{table}[t]
\scalebox{0.87}{
\begin{tabular}{l|llllll|l}
\hline
              & B4   & M    & R    & C     & S    & BS    & ACC   \\ \hline
ReRe          & 29.2 & 23.4 & 52.1 & 113.4 & 22.7 & 90.2  & 83.7  \\ 
$Oracle_{e}$     & \textbf{36.4} & \textbf{27.9} & \textbf{58.2} & \textbf{142.1} & \textbf{27.1} & \textbf{90.85} & 83.84 \\
$Oracle_{ae}$     & 30.8 & 24.3 & 52.6 & 118.9 & 24.0 & 90.43 & \textbf{94.10} \\ \hline
\end{tabular}
}
\caption{Filtered score of oracle test.
$Oracle_{e}$ using only answer feature and $Oracle_{ae}$ use  answer and explanation features.
$Oracle_{e}$ and $Oracle_{ae}$ show outstanding explanation score and accuracy in line with our intuition.}
\vspace{-0.1cm}
\label{tab:my-table}
\end{table}

\subsection {Automatic Evaluation}
In Table 1, we present the performance scores compared to state-of-the-art models in the filtered version.
The scoring method includes unfiltered scores, which measure all predictions regardless of whether they are correct or not, and filtered scores, which measure only the predictions that match the correct answers. 
In VQA-NLE, generating a good explanation based on accurate answers is important, and providing a good explanation for incorrect answers is meaningless.
Therefore, filtered scores are given more consideration.
Follow the \cite{acc1,sammani2022nlx}, VQA accuracy is measured as correct when the predicted answers are within the expected answers. Experimental results show that measuring the similarity between the query's image and the sample's explanation shows higher performance than measuring image-image similarity for memory retrieval.
Compared to recent state-of-the-art models, OURS shows a performance improvement of 2$\sim$3\% in the metric measured by explanation score.
Through these results, we can confirm that retrieval information helps generate more accurate answers and higher-quality explanations.

\begin{figure}
    \centering
    \centerline{\includegraphics[width=0.5\textwidth]{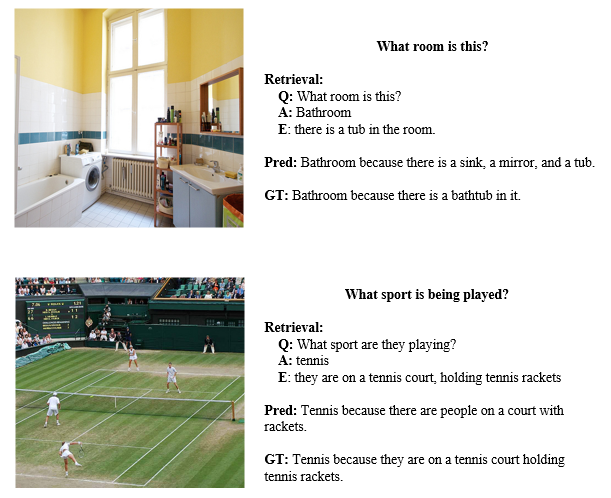}}
    \vspace{-0.2cm}
    \caption{Example of retrieval sample and model generated prediction compare to ground truth.}
\end{figure}

\subsection{Oracle Test}

From Table 2, we can see how much the performance of our model can be improved when the ideal retrieval samples are retrieved.
Ideal retrieval is retrieved from a memory database, using cosine similarity of ground truth answer, explanation with sample's answer, and explanation.

We conducted an Oracle test considering two cases: one using only the answer feature for input retrieval features and the other using both the answer and explanation features.
The result of the Oracle test shows that in line with our intuition, the explanation score significantly improved with the aid of the ideal sample's explanation feature.
When the ideal answer feature is given with the explanation feature, Accuracy is raised to 94.10.
These results demonstrate that a simple structure of adding a cross-attention block in the LM block is sufficient for the model to gain reasoning from retrieval features.

\section{Conclusion}

In this paper, we propose applying the retrieval augmentation method to the VQA-NLE task.
We define appropriate retrieval in terms of question type and semantic consistency.
For retrieving appropriate samples, we utilize cosine similarity on feature combinations.
The retrieved features are processed by the cross-attention in the GPT-2 language model.
Through these processes, ReRe generates answers and explanations simultaneously to aid with the retrieval feature.
ReRe shows improvement in accuracy and explanation score on VQA-X. It will be interesting future work to explore larger memory with better similarity matching to further improve the performance of the VQA-NLE task.


\vfill\pagebreak
\label{sec:refs}

\bibliographystyle{IEEEbib}
\bibliography{refs}

\begin{thebibliography}{10}

\bibitem{anderson2018vision}
Peter Anderson, Qi~Wu, Damien Teney, Jake Bruce, Mark Johnson, Niko S{\"u}nderhauf, Ian Reid, Stephen Gould, and Anton Van Den~Hengel,
\newblock ``Vision-and-language navigation: Interpreting visually-grounded navigation instructions in real environments,''
\newblock in {\em Proceedings of the IEEE conference on computer vision and pattern recognition}, 2018, pp. 3674--3683.

\bibitem{feng2023nle}
Yuchao Feng, Wei Hua, and Yuxiang Sun,
\newblock ``Nle-dm: Natural-language explanations for decision making of autonomous driving based on semantic scene understanding,''
\newblock {\em IEEE Transactions on Intelligent Transportation Systems}, 2023.

\bibitem{RATIONALETRANSFORMER}
Ana Marasovi{\'c}, Chandra Bhagavatula, Jae~Sung Park, Ronan~Le Bras, Noah~A Smith, and Yejin Choi,
\newblock ``Natural language rationales with full-stack visual reasoning: From pixels to semantic frames to commonsense graphs,''
\newblock {\em Findings of EMNLP}, 2020.

\bibitem{lai2024towards}
Chengen Lai, Shengli Song, Shiqi Meng, Jingyang Li, Sitong Yan, and Guangneng Hu,
\newblock ``Towards more faithful natural language explanation using multi-level contrastive learning in vqa,''
\newblock in {\em Proceedings of the AAAI Conference on Artificial Intelligence}, 2024, vol.~38, pp. 2849--2857.

\bibitem{suo2023s3c}
Wei Suo, Mengyang Sun, Weisong Liu, Yiqi Gao, Peng Wang, Yanning Zhang, and Qi~Wu,
\newblock ``S3c: Semi-supervised vqa natural language explanation via self-critical learning,''
\newblock in {\em Proceedings of the IEEE/CVF Conference on Computer Vision and Pattern Recognition}, 2023, pp. 2646--2656.

\bibitem{ge2023wrong}
Jiaxin Ge, Sanjay Subramanian, Trevor Darrell, and Boyi Li,
\newblock ``From wrong to right: A recursive approach towards vision-language explanation,''
\newblock {\em Empirical Methods in Natural Language Processing (EMNLP)}, 2023.

\bibitem{lu2022learn}
Pan Lu, Swaroop Mishra, Tanglin Xia, Liang Qiu, Kai-Wei Chang, Song-Chun Zhu, Oyvind Tafjord, Peter Clark, and Ashwin Kalyan,
\newblock ``Learn to explain: Multimodal reasoning via thought chains for science question answering,''
\newblock {\em Advances in Neural Information Processing Systems}, vol. 35, pp. 2507--2521, 2022.

\bibitem{pan2023retrieving}
Junting Pan, Ziyi Lin, Yuying Ge, Xiatian Zhu, Renrui Zhang, Yi~Wang, Yu~Qiao, and Hongsheng Li,
\newblock ``Retrieving-to-answer: Zero-shot video question answering with frozen large language models,''
\newblock in {\em Proceedings of the IEEE/CVF International Conference on Computer Vision}, 2023, pp. 272--283.

\bibitem{ramos2023smallcap}
Rita Ramos, Bruno Martins, Desmond Elliott, and Yova Kementchedjhieva,
\newblock ``Smallcap: lightweight image captioning prompted with retrieval augmentation,''
\newblock in {\em Proceedings of the IEEE/CVF Conference on Computer Vision and Pattern Recognition}, 2023, pp. 2840--2849.

\bibitem{sarto2022retrieval}
Sara Sarto, Marcella Cornia, Lorenzo Baraldi, and Rita Cucchiara,
\newblock ``Retrieval-augmented transformer for image captioning,''
\newblock in {\em Proceedings of the 19th international conference on content-based multimedia indexing}, 2022, pp. 1--7.

\bibitem{firstVQA}
Mateusz Malinowski and Mario Fritz,
\newblock ``A multi-world approach to question answering about real-world scenes based on uncertain input,''
\newblock {\em Advances in neural information processing systems}, vol. 27, 2014.

\bibitem{vqa1}
Peter Anderson, Xiaodong He, Chris Buehler, Damien Teney, Mark Johnson, Stephen Gould, and Lei Zhang,
\newblock ``Bottom-up and top-down attention for image captioning and visual question answering,''
\newblock in {\em Proceedings of the IEEE conference on computer vision and pattern recognition}, 2018, pp. 6077--6086.

\bibitem{vqa3}
Chao Ma, Chunhua Shen, Anthony Dick, Qi~Wu, Peng Wang, Anton Van~den Hengel, and Ian Reid,
\newblock ``Visual question answering with memory-augmented networks,''
\newblock in {\em Proceedings of the IEEE conference on computer vision and pattern recognition}, 2018, pp. 6975--6984.

\bibitem{first-vqa-nle}
Lisa~Anne Hendricks, Zeynep Akata, Marcus Rohrbach, Jeff Donahue, Bernt Schiele, and Trevor Darrell,
\newblock ``Generating visual explanations,''
\newblock in {\em ECCV}, 2016, pp. 3--19.

\bibitem{gradcam}
Ramprasaath~R Selvaraju, Michael Cogswell, Abhishek Das, Ramakrishna Vedantam, Devi Parikh, and Dhruv Batra,
\newblock ``Grad-cam: Visual explanations from deep networks via gradient-based localization,''
\newblock in {\em CVPR}, 2017, pp. 618--626.

\bibitem{vqax}
Dong~Huk Park, Lisa~Anne Hendricks, Zeynep Akata, Anna Rohrbach, Bernt Schiele, Trevor Darrell, and Marcus Rohrbach,
\newblock ``Multimodal explanations: Justifying decisions and pointing to the evidence,''
\newblock in {\em Proceedings of the IEEE conference on computer vision and pattern recognition}, 2018, pp. 8779--8788.

\bibitem{e-ug}
Maxime Kayser, Oana-Maria Camburu, Leonard Salewski, Cornelius Emde, Virginie Do, Zeynep Akata, and Thomas Lukasiewicz,
\newblock ``e-vil: A dataset and benchmark for natural language explanations in vision-language tasks,''
\newblock in {\em ICCV}, 2021, pp. 1244--1254.

\bibitem{uniter}
Yen-Chun Chen, Linjie Li, Licheng Yu, Ahmed El~Kholy, Faisal Ahmed, Zhe Gan, Yu~Cheng, and Jingjing Liu,
\newblock ``Uniter: Universal image-text representation learning,''
\newblock in {\em ECCV}, 2020, pp. 104--120.

\bibitem{gpt2}
Alec Radford, Jeffrey Wu, Rewon Child, David Luan, Dario Amodei, Ilya Sutskever, et~al.,
\newblock ``Language models are unsupervised multitask learners,''
\newblock {\em OpenAI blog}, vol. 1, no. 8, pp. 9, 2019.

\bibitem{sammani2022nlx}
Fawaz Sammani, Tanmoy Mukherjee, and Nikos Deligiannis,
\newblock ``Nlx-gpt: A model for natural language explanations in vision and vision-language tasks,''
\newblock in {\em proceedings of the IEEE/CVF conference on computer vision and pattern recognition}, 2022, pp. 8322--8332.

\bibitem{whitehouse2023towards}
Chenxi Whitehouse, Tillman Weyde, and Pranava Madhyastha,
\newblock ``Towards a unified model for generating answers and explanations in visual question answering,''
\newblock {\em arXiv preprint arXiv:2301.10799}, 2023.

\bibitem{retrieval1}
Sebastian Borgeaud, Arthur Mensch, Jordan Hoffmann, Trevor Cai, Eliza Rutherford, Katie Millican, George~Bm Van Den~Driessche, Jean-Baptiste Lespiau, Bogdan Damoc, Aidan Clark, et~al.,
\newblock ``Improving language models by retrieving from trillions of tokens,''
\newblock in {\em International conference on machine learning}. PMLR, 2022, pp. 2206--2240.

\bibitem{retrieval2}
Kelvin Guu, Kenton Lee, Zora Tung, Panupong Pasupat, and Mingwei Chang,
\newblock ``Retrieval augmented language model pre-training,''
\newblock in {\em International conference on machine learning}. PMLR, 2020, pp. 3929--3938.

\bibitem{retrieval4}
Patrick Lewis, Ethan Perez, Aleksandra Piktus, Fabio Petroni, Vladimir Karpukhin, Naman Goyal, Heinrich K{\"u}ttler, Mike Lewis, Wen-tau Yih, Tim Rockt{\"a}schel, et~al.,
\newblock ``Retrieval-augmented generation for knowledge-intensive nlp tasks,''
\newblock {\em Advances in Neural Information Processing Systems}, vol. 33, pp. 9459--9474, 2020.

\bibitem{kim2024you}
Minkuk Kim, Hyeon~Bae Kim, Jinyoung Moon, Jinwoo Choi, and Seong~Tae Kim,
\newblock ``Do you remember? dense video captioning with cross-modal memory retrieval,''
\newblock {\em arXiv preprint arXiv:2404.07610}, 2024.

\bibitem{radford2021clip}
Alec Radford, Jong~Wook Kim, Chris Hallacy, Aditya Ramesh, Gabriel Goh, Sandhini Agarwal, Girish Sastry, Amanda Askell, Pamela Mishkin, Jack Clark, et~al.,
\newblock ``Learning transferable visual models from natural language supervision,''
\newblock in {\em International conference on machine learning}. PMLR, 2021, pp. 8748--8763.

\bibitem{papineni2002bleu}
Kishore Papineni, Salim Roukos, Todd Ward, and Wei-Jing Zhu,
\newblock ``Bleu: a method for automatic evaluation of machine translation,''
\newblock in {\em ACL}, 2002, pp. 311--318.

\bibitem{banerjee2005meteor}
Satanjeev Banerjee and Alon Lavie,
\newblock ``Meteor: An automatic metric for mt evaluation with improved correlation with human judgments,''
\newblock in {\em Proceedings of the acl workshop on intrinsic and extrinsic evaluation measures for machine translation and/or summarization}, 2005, pp. 65--72.

\bibitem{lin2004rouge}
Chin-Yew Lin,
\newblock ``Rouge: A package for automatic evaluation of summaries,''
\newblock in {\em Text summarization branches out}, 2004, pp. 74--81.

\bibitem{vedantam2015cider}
Ramakrishna Vedantam, C~Lawrence~Zitnick, and Devi Parikh,
\newblock ``Cider: Consensus-based image description evaluation,''
\newblock in {\em CVPR}, 2015, pp. 4566--4575.

\bibitem{anderson2016spice}
Peter Anderson, Basura Fernando, Mark Johnson, and Stephen Gould,
\newblock ``Spice: Semantic propositional image caption evaluation,''
\newblock in {\em ECCV}, 2016, pp. 382--398.

\bibitem{zhang2019bertscore}
Tianyi Zhang, Varsha Kishore, Felix Wu, Kilian~Q Weinberger, and Yoav Artzi,
\newblock ``Bertscore: Evaluating text generation with bert,''
\newblock {\em ICLR}, 2020.

\bibitem{FME}
Jialin Wu and Raymond~J Mooney,
\newblock ``Faithful multimodal explanation for visual question answering,''
\newblock {\em ACL BlackboxNLP workshop}, 2019.

\bibitem{acc1}
Qian Yang, Yunxin Li, Baotian Hu, Lin Ma, Yuxin Ding, and Min Zhang,
\newblock ``Chunk-aware alignment and lexical constraint for visual entailment with natural language explanations,''
\newblock in {\em Proceedings of the 30th ACM International Conference on Multimedia}, 2022, pp. 3587--3597.

\end{thebibliography}

\end{document}